\documentclass[sigconf,nonacm]{acmart}

\AtBeginDocument{%
  \providecommand\BibTeX{{%
    \normalfont B\kern-0.5em{\scshape i\kern-0.25em b}\kern-0.8em\TeX}}}

\acmConference[Preprint]{Preprint}
\usepackage{comment}
\usepackage{algorithm}
\usepackage[english]{babel}
\usepackage{hyperref}
\usepackage[noend]{algpseudocode}
\begin{document}

\title{IERL: \underline{I}nterpretable \underline{E}nsemble \underline{R}epresentation \underline{L}earning - Combining CrowdSourced Knowledge and Distributed Semantic Representations}

\author{Yuxin Zi}
\email{yzi@email.sc.edu}
\affiliation{%
  \institution{Artificial Intelligence Institute \\ University of South Carolina}
  \country{USA}
}
\author{Kaushik Roy}
\email{kaushikr@email.sc.edu}
\affiliation{%
  \institution{Artificial Intelligence Institute \\ University of South Carolina}
  \country{USA}
}

\author{Vignesh Narayanan}
\email{vignar@sc.edu}
\affiliation{%
  \institution{Artificial Intelligence Institute \\ University of South Carolina}
  \country{USA}
}
\author{Manas Gaur}
\email{manas@umbc.edu}
\affiliation{%
  \institution{KAI\textsuperscript{2}, University of Maryland \\ Baltimore County}
  \country{USA}
}
\author{Amit Sheth}
\email{amit@sc.edu}
\affiliation{%
  \institution{Artificial Intelligence Institute \\ University of South Carolina}
  \country{USA}
}

\begin{abstract}
Large Language Models (LLMs) encode meanings of words in the form of distributed semantics. Distributed semantics capture common statistical patterns among language tokens (words, phrases, and sentences) from large amounts of data. LLMs perform exceedingly well across General Language Understanding Evaluation (GLUE) tasks designed to test a model's understanding of the meanings of the input tokens. 
However, recent studies have shown that LLMs tend to generate unintended, inconsistent, or wrong texts as outputs when processing inputs that were seen rarely during training, or inputs that are associated with diverse contexts (e.g., well-known \emph{hallucination} phenomenon in language generation tasks). Crowdsourced and expert-curated knowledge graphs such as ConceptNet are designed to capture the meaning of words from a compact set of well-defined contexts. Thus LLMs may benefit from leveraging such knowledge contexts to reduce inconsistencies in outputs. We propose a novel ensemble learning method, the \textit{Interpretable Ensemble Representation Learning} (IERL), that systematically combines LLM and crowdsourced knowledge representations of input tokens. IERL has the distinct advantage of being interpretable by design (when was the \textit{LLM context} used vs. when was the \textit{knowledge context} used?) over state-of-the-art (SOTA) methods, allowing scrutiny of the inputs in conjunction with the 
parameters of the model, facilitating the analysis of 
models' inconsistent or irrelevant outputs. Although IERL is agnostic to the choice of LLM and crowdsourced knowledge, we demonstrate our approach using BERT and ConceptNet. We 
report improved or competitive results 
with IERL across GLUE tasks over current SOTA methods and significantly enhanced model interpretability. 
\end{abstract}





\maketitle
\section{Introduction}\label{introduction}
LLMs have performed exceedingly well on the GLUE benchmark tasks \cite{wei2022emergent}. GLUE tasks measure the machine's comprehension on supervised learning-based natural language processing tasks, such as Quora Question Pairs to check question redundancy, and Recognizing Textual Entailment to check if two sentences share entailment, neutral, or contraction relations \cite{wang2018glue}.
LLMs learn trillions of parameters after training over a humongous amount of data. Irregularities in the data (for example, little or highly varying language token patterns or contexts) causes 
LLMs to \textit{hallucinate} - generating inconsistent outputs for similar inputs. 

Crowdsourced and curated knowledge graphs (KGs), such as ConceptNet, are designed to capture meanings of commonly used words using a compact set of contexts agreed by humans \cite{bricker2014ontological,speer2017conceptnet}. As a result, representations learned from ConceptNet are less likely to suffer from distributional irregularities among the tokens. Consequently, it is a promising and an active research topic on utilizing representations from KGs to potentially mitigate irregularities, while processing input tokens via LLM representations. In this paper, we focus on developing a learning method that systematically incorporates representations from both KGs and LLMs to address the following unresolved questions. \textbf{Q1}: \textit{Can we design an approach to combine crowdsourced knowledge and LLM representations to obtain an integrated representation, in order to mitigate the model hallucination?} \textbf{Q2}: \textit{Can we achieve an interpretable design - i.e., can we tractably discern for what inputs the LLM hallucinates on and what knowledge context improves representation quality?} Next, we briefly review existing methods that seek to infuse representations from KGs and LLMs and their relevance to questions \textbf{Q1} and \textbf{Q2}.

\subsection{Related Work on Combining Knowledge and LLM Representations}
There is an extensive literature on combining LLMs and knowledge representations to leverage contextual information among language tokens from both \cite{ji2021survey}. The representations are then processed through a task-specific neural network. Here we will cover the four SOTA approaches, KALA, K-Adapter, TDLR, and GCT, broadly representing two kinds of methods - (1) Combining representations at the input level before passing it through the neural network (KALA and K-Adapter) and (2) Combing representations at the parametric level, i.e., modify the parameters of the task-specific neural network and the resulting token meaning interpretations (TDLR and GCT) \cite{kang2022kala,wang2020k,rawte22tdlr,choi2020learning}.

KALA modifies the input LLM representations for tokens by using a weighted aggregate of other tokens connected in the knowledge graph. K-Adapter trains ``adapter'' models for encoding knowledge representations and combines the LLM and adapter representations at the input level. With KALA and K-Adapter, it is not possible to keep track of and understand how the representations are incorporated into the neural network after the input stage internally. Ablation studies and post-hoc approximate interpretations using LIME, etc., provide representation interpretability (was the knowledge context important or the data context?)\cite{mishra2017local}. However, it is unclear how far the approximation is off from the truth, which is crucial to evaluating and systematically addressing LLM hallucination issues.
Furthermore, the effect of KALA and K-Adapter representations on hallucinations has not been studied. Thus, KALA and K-Adapter do not fully address \textbf{Q1} or \textbf{Q2}. TDLR operates on the self-attention mechanism of transformers by modifying the attention or weight matrices to hard-code graph connections among language tokens. TDLR does this once, in the first self-attention block, and then allows model fine-tuning to continue as is. It is unclear if the attention matrix modification is retained during the fine-tuning across the remaining transformer blocks. GCT is similar to TDLR and differs in the specifics of the self-attention matrix modification operation. TDLR and GCT suffer from similar issues as KALA and K-Adapter towards addressing \textbf{Q1} and \textbf{Q2}. 

\section{Background and Motivation}\label{background}
In this section, we describe the GLUE tasks, what hallucination looks like when solving the GLUE tasks and the theoretical motivations for the IERL algorithm presented in Section \ref{algorithm}.
\subsection{Task Descriptions and Hallucinations}\label{tasks}
We experiment with similarity or entailment GLUE tasks that take a pair of sentences as input and format its output as a $+1$ or a $-1$. For the similarity tasks, $+1$ and $-1$ correspond to similar or dissimilar input sentences respectively. For the entailment tasks, +1 and -1 correspond to entailment and contradiction, respectively. 

We use $X$ to denote the dataset, and $x$ is an instance in the dataset $X$. Each $x$ is a three-tuple composed of $x[1]$: sentence 1, $x[2]$: sentence 2, and the label $y$. 
\paragraph{\textbf{Hallucinations.}} Hallucinations refers to inconsistent model outputs for similar inputs resulting from statistical irregularities in the data. Since this notion is often used in the context of language generation tasks, we clarify the context in which we use it in this paper. To formalize this notion, we batch our instances into random batches. We note the convergence rate variations of the training loop across these batches, where each batch is of size equal to 80\% of the training dataset $X$. We find that for the GLUE tasks, SOTA fine-tuning results in a high convergence rate variance (ranges from $13-45$ iterations). This suggests that there may be a high degree of irregularity in the statistical properties across the batches. Therefore, we can expect a model trained on these datasets to generate inconsistent outputs for similar inputs, i.e., suffer from hallucinations.
\subsection{Ensemble Learning Approach}\label{approach}
To fine-tune the models for GLUE tasks, a few feedforward neural network layers are added and trained using backpropagation. Such a training procedure works well when there is a generalizable pattern across the instances in the fine-tuning dataset (regularly occurring statistical patterns). To tackle the issue of irregularities, we propose using example patterns from each instance and aggregating them using an ensemble learning approach. We can think of an example pattern as one that maps a given instance to its output in the task-specific dataset, which can be seen as (an instance level)  model and define an ensemble function $g(z)$ for a new point instance $z$ as an ensemble of weighted contributions from similar instances in the dataset $X$ as
\begin{align}
    g(z) = \sum_{\color{red}x}(\alpha^1_{x[1]}\odot \langle z,x[1] \rangle + \alpha^2_{x[2]}\odot \langle z,x[2] \rangle).
    \label{eqn:ensemble}
\end{align}
Here $\odot$ refers to a product operation defined in the algorithm and experimentation sections, and $\langle\cdot\rangle$ refers to a suitable similarity computation. 
\subsection{Utilizing Knowledge Graph Contexts}
The ensemble approach formulation allows the expressiveness to model both generalizability across instances and instance-level details. Examining the ensemble model's parameters lets us interpret whether an instance shows irregular patterns. However, it does not yet incorporate a mechanism to solve the irregularity issue during model learning. Here we posit that combining LLM and knowledge graph representations using an ensemble approach allows us to interpret instances for their pattern regularities and draw from either the LLM or knowledge contexts to solve the irregularity resulting in high performance that hallucinates less.

Thus, we expand the formulation in \eqref{eqn:ensemble} to describe the operation of combining LLM and knowledge representations as
\begin{equation}
    \scriptsize
    g(z) = \sum_{x}(\alpha_{LLM}^{sim} \cdot [\langle z,x[1]\rangle_{LLM}^{sim}~\langle z,x[2]\rangle_{LLM}^{sim}] + \alpha_{KG}^{sim} \cdot [\langle z,x[1]\rangle_{KG}^{sim}~\langle z,x[2]\rangle_{KG}^{sim}])
    \label{eqn:ensemble2}
\end{equation}
Here, $\cdot$ refers to the dot product between vectors, $\langle\cdot\rangle_{LLM}^{sim}$ refers to a similarity measure between the LLM representations, and $\langle\cdot\rangle_{KG}^{sim}$ refers to a  similarity measure between the KG embedding representations. The $\alpha_{LLM}^{sim}$ and $\alpha_{KG}^{sim}$ are two dimensional vectors.

\paragraph{\textbf{Aggregation Methods}}
Aggregation as a method to reduce statistical irregularities such as high variance has been well-studied in statistical learning theory literature \cite{clark1976effects}. The ensemble formulation in \eqref{eqn:ensemble2} can be seen as aggregation over instances in the dataset $X$. In this work, we experiment with two types of aggregation, the average of the instance representations and using averages over higher-order moment representations. We can expect LLMs trained on very large amounts of data to tend to the \textit{normal} distributional trend in the underlying data distribution. Therefore the average (first-order moment) and variance (second-order moment) of groups of instance representations are \textit{sufficient statistics} to describe the underlying distribution. However, for smaller number of data instances (such as in GLUE task datasets), it may be necessary to utilize averages over higher order moments as \textit{sufficient statistics}. We experiment with both types of aggregation and compare the results.

\section{The Interpretable Ensemble Representation Learning (IERL) Algorithm}\label{algorithm}
\begin{figure}[!h]
    \centering
    \includegraphics[width=\linewidth,keepaspectratio]{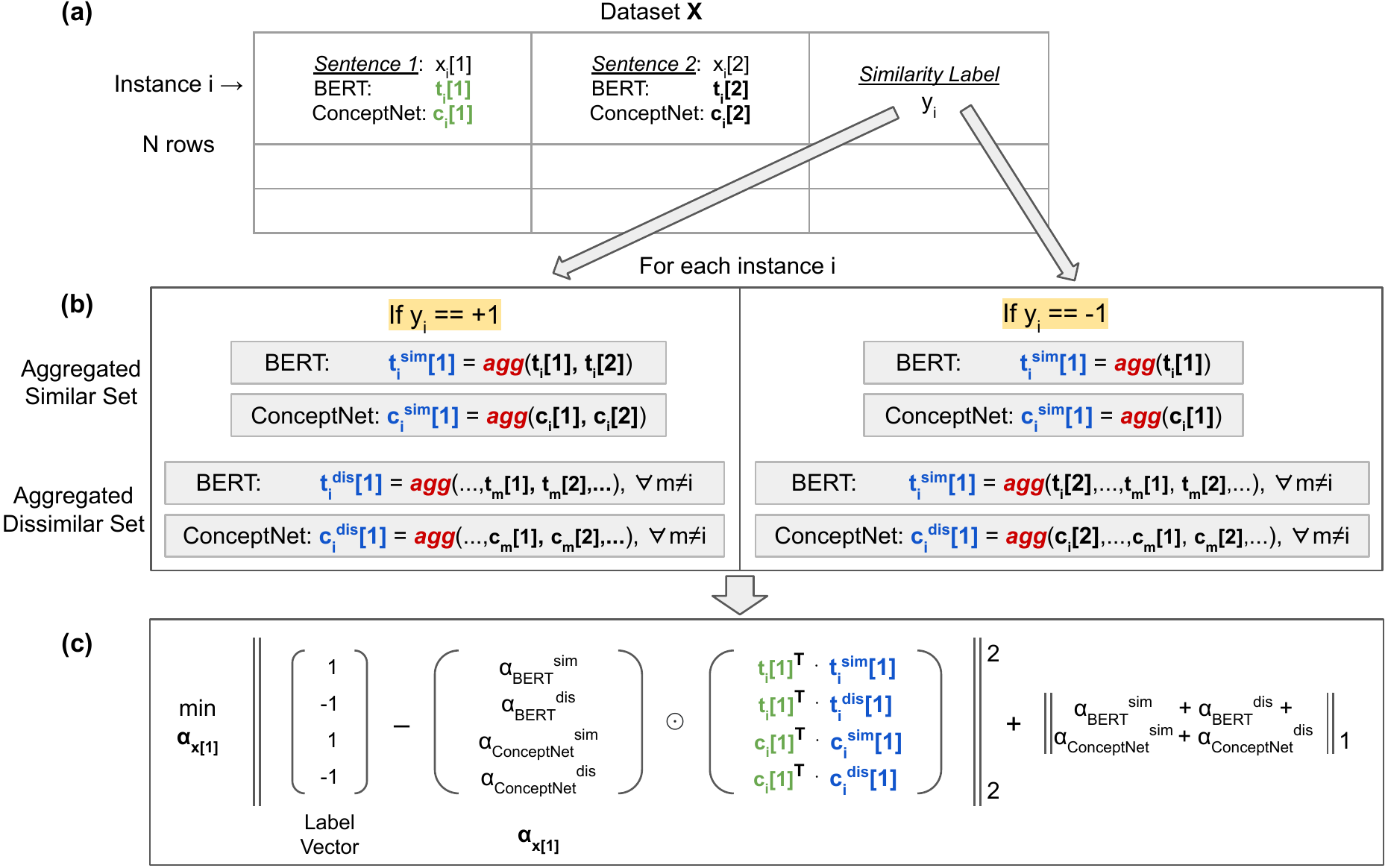}
    \includegraphics[width=\linewidth,keepaspectratio]{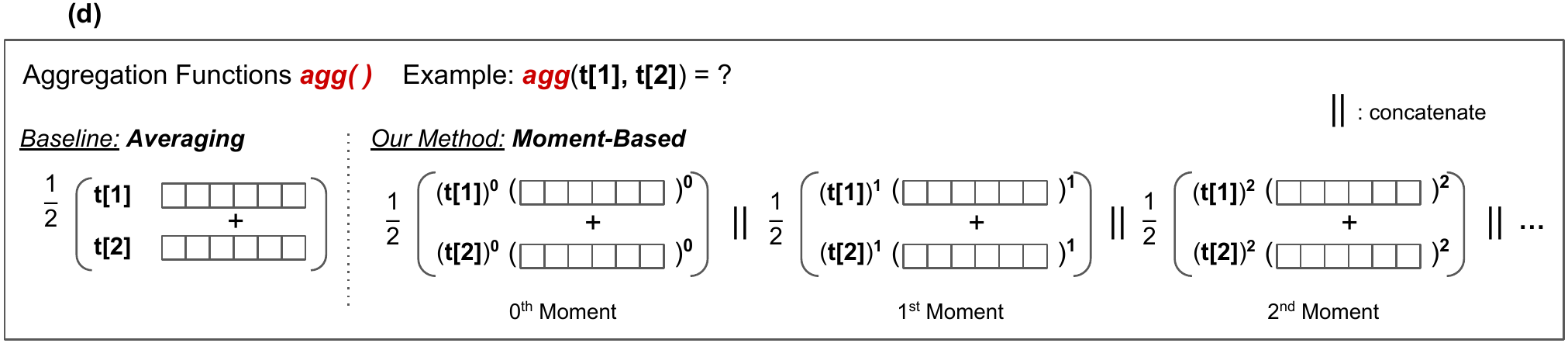}
    \caption{\textbf{(a)} Shows the dataset $X$ (e.g., Recognizing Textual Entailment (RTE)) and its instances $x_i$ indexed by $i$ ($i=1,\ldots, N$). The BERT representations of sentence 1: $x_i[1]$ and 2: $x_i[2]$ from instance $i$ are denoted by $t_i[1]$ and $t_i[2]$. The ConceptNet representations of sentences 1 and 2 from instance $i$ are denoted by $c_i[1]$ and $c_i[2]$, respectively. \textbf{(b)} Shows how similar and dissimilar instances to $x_i[1]$ are constructed and aggregated for the cases of $y_i = 1$ and $-1$, respectively. \textbf{(c)} Shows one of the optimization problems in detail corresponding to line 22 in algorithm \ref{algorithm}. \textbf{(d)} Shows two methods of aggregation over instances - Averaging and Moment-Based (algorithm \ref{agg}) aggregation.}
    \label{fig:alg_overview}
\end{figure}

Figure \ref{fig:alg_overview} shows an illustration of the IERL optimization step - \textbf{(a)} Shows the dataset $X$ (e.g., Recognizing Textual Entailment) and its instances $x_i$ indexed by $i$. $t_i[1]$ and $t_i[2]$ denote the BERT representations of sentence 1: $x_i[1]$ and 2: $x_i[2]$ from instance $i$. $c_i[1]$ and $c_i[2]$ denote the ConceptNet representations of sentences 1 and 2 from instance $i$. \textbf{(b)} Shows how similar and dissimilar instances to $x_i[1]$ are constructed and aggregated for the cases of $y_i == 1$ and $-1$ respectively. \textbf{(c)} Shows one step of optimization in detail corresponding to line 22 in Algorithm \ref{algorithm} \textbf{(d)} Shows two methods of aggregation over instances - Averaging and Moment-Based (Algorithm \ref{agg}) aggregation.
Algorithm \ref{IERL} and \ref{agg} detail the IERL and aggregation algorithm, respectively.
\begin{algorithm}[!h]
\caption{Interpretable Ensemble Representation Learning (IERL)} \label{IERL}
\begin{algorithmic}[1]
    \State \textbf{Inputs:} Dataset: $x \in X$ \Comment $x = (sentence1, sentence2, label)$, see section \ref{tasks}
    \State Models = $\{~\}$ \Comment Initialize dictionary to store models for all sentences in $X$
    \For{$i \in X$} \Comment i indexes instance $x$
    \State $x_i = X[i]$ \Comment ith instance
    \State $t_i[1],~t_i[2] = LLM(x_i[1]),~LLM(x_i[2])$ \Comment LLM representations for sentences
    \State $c_i[1],~c_i[2] = KG(x_i[1]),~KG(x_i[2])$ \Comment KG representations for sentences
    \State $y = x[3]$ \Comment Label for this instance
    \For{$j \in \{1,2\}$} \Comment j indexes each sentence in the instance
    \If{y == +1} \Comment Similar and dissimilar instance aggregation
    \State $t_i^{sim}[j] = ${\color{red}agg}$([t_i[1],~t_i[2]])$ 
    \State $c_i^{sim}[j] = ${\color{red}agg}$([c_i[1],~c_i[2]])$
    \State $t_i^{dis}[j] = ${\color{red}agg}$([..,t_m[1],t_m[2],..]), \forall m \neq i$ 
    \State $c_i^{dis}[j] = ${\color{red}agg}$([..,c_m[1],c_m[2],..]), \forall m \neq i$
    \ElsIf{y == -1} \Comment Similar and dissimilar instance aggregation
    \State $t_i^{sim}[j] = ${\color{red}agg}$([t_i[1]])$ 
    \State $c_i^{sim}[j] = ${\color{red}agg}$([c_i[1]])$
    \State $t_i^{dis}[j] = ${\color{red}agg}$([t_i[2],..,t_m[1],t_m[2],..]), \forall m \neq i$ 
    \State $c_i^{dis}[j] = ${\color{red}agg}$([c_i[2],..,c_m[1],c_m[2],..]), \forall m \neq i$
    \EndIf
    \State $\alpha_i[j] = [\alpha_{LLM}^{sim}, \alpha_{LLM}^{dis}, \alpha_{KG}^{sim}, \alpha_{KG}^{dis}]$ \Comment parameters for instance $i$, sentence $j$
    \State $I = [1, -1, 1,-1]$
    \State $D = [t_i[j]\cdot t_i^{sim}[j], t_i[j]\cdot t_i^{dis}[j], c_i[j]\cdot c_i^{sim}[j], c_i[j]\cdot c_i^{dis}[j]]$
    \State Optimization until convergence: Minimize $g_i[j] = ||I - \alpha_i[j]\odot D||_2^2 + ||\alpha_i[j]||_1^1$
    \State Models$[x_i[j]] = (g_i[j],\alpha_i[j])$ \Comment Store model for instance $i$, sentence $j$
    \EndFor
    \EndFor
    \State return Models
\end{algorithmic}
\end{algorithm}
\begin{algorithm}[!h]
\caption{Aggregation Algorithm ({\color{red}agg})} \label{agg}
\begin{algorithmic}[1]
    \State \textbf{Inputs:} list of vectors $V$
    \State $agg_V = [~]$
    \For{$v \in V$} \Comment calculate element wise powers of the vector elements
    \State $v_0,v_1,v_2,v_3 = v^0,v^1,v^2,v^3$ 
    \State $v_{concat} = concat(v_0,v_1,v_2,v_3)$ \Comment Concatenate the power vectors
    \State $agg_V.append(v_{concat})$ \Comment add to list of vectors to aggregate
    \EndFor
    \State return $mean(agg_V)$ \Comment return average of all lists in $agg_V$
\end{algorithmic}
\end{algorithm}
\section{Experimental Section and Results}\label{experiments}
In our experiments, we use BERT as the choice of $LLM$ representations and ConceptNet NumberBatch embeddings for the choice of $KG$ representations in the IERL algorithm (\ref{algorithm}). We use gradient descent as our optimization procedure, and use grid search to tune hyperparameters for optimization. For the computation of higher order moment representations in algorithm \ref{agg}, we execute each for loop iteration in parallel. We initialize the parameters $\alpha_i[j]$ for each $(i,j)$ using a $0$ mean, $I$ covariance 4d-gaussian distribution. We test our method on the GLUE tasks: Quora Question Pairs (QQP), Question-Answering NLI (QNLI), Multi-Genre NLI (MNLI), RTE, Stanford Sentiment Treebank v2 (SST-2), and Winograd NLI (WNLI) pertaining to two types of tasks:
\begin{enumerate}
    \item \textbf{Sentence Similarity}: Consists of input sentence pairs and a $1$ or $0$ denoting if the pairs are similar or not (we reformulate to $1$ and $-1$) - QQP, and STS
    \item \textbf{Sentence Entailment}: Consists of input sentence pairs and label from among ``entailment, contradiction, neutral'' (we reformulate to $1$ for entailment and $-1$ for contradiction) - QNLI, WNLI, MNLI, and RTE
\end{enumerate}
We also convert our vectors to unit vectors before computing dot products (line 21 in IERL Algorithm - \ref{algorithm}).
\paragraph{\textbf{Baseline Model}} For our baseline model we implement IERL using a simple average for aggregation (instead of computing moments using Algorithm \ref{agg}). We call this IERL\textsubscript{B}. We present our evaluation results in the order that they address the questions \textbf{Q1} and \textbf{Q2} introduced in section \ref{introduction}.
\subsection{Quantitative Evaluation - Addresses \textbf{Q1}}\label{quant}
We report accuracy measures of our method against the baseline IERL\textsubscript{B} and the current leader on the GLUE leaderboard and see that the performance of IERL shows competitive performance even against state-of-the-art performance. We also compute \#Optimization steps using randomly sampled batches of size 80\% of the whole dataset per sample and tabulate the range (min-max)\footnote{We are currently running fine-tuning using the leaderboard model and will report \#Optimization-Steps range in future work}. We see that the range is significantly higher using an implementation of BERT (BERT (Ours) - Vanilla BERT with 6 layers and fine-tuning) compared to both versions of IERL. Furthermore, higher-order moments also show a much faster convergence of 7-13 steps vs. 20-30 and 20-45 steps. We use up to fourth-order moments in our experiments, i.e., 0-3.
\begin{table}[!h] 
\begin{tabular}{p{0.7cm}|p{0.6cm}|p{0.6cm}|p{0.6cm}|p{0.6cm}|p{0.6cm}|p{0.6cm}|p{0.7cm}}
\hline
System                     & STS        & QQP  & QNLI & WNLI & MNLI & RTE  & \#Opt    \\ \hline
GLUE     & 93.5         & 90.9          & 96.7          & 97.9          & 92.5   & 93.6 & -       \\
BERT     & 89.7         & 88.7          & 93.5          & 93.3          & 81.5   & 88.3   & 20-45     \\
IERL\textsubscript{B}     & 90.89          & 86.41          & 92.3          & 90.11          & 88.53 &  90.4 & 20-30          \\ 
\textbf{IERL}   & \textbf{93.55} & \textbf{90.51} & \textbf{95.56} & \textbf{98.7}  & \textbf{92.08} & \textbf{92.3} & \textbf{7-13}\\ \hline
\end{tabular}
\caption{Comparing IERL performance on similarity and entailment GLUE tasks. We also see that the \# of Optimization steps (\#Opt) stabilizes using the IERL training method.  IERL shows competitive performance even against state-of-the-art performance. Using higher-order moments in IERL shows a much faster convergence of 7-13 steps vs. 20-45 and 20-30 steps.}
\label{tab:quant}
\end{table}
\subsection{Qualitative Evaluation - Addresses \textbf{Q2}}\label{qual}
\begin{figure}[!h]
    \centering
    \includegraphics[width=\linewidth]{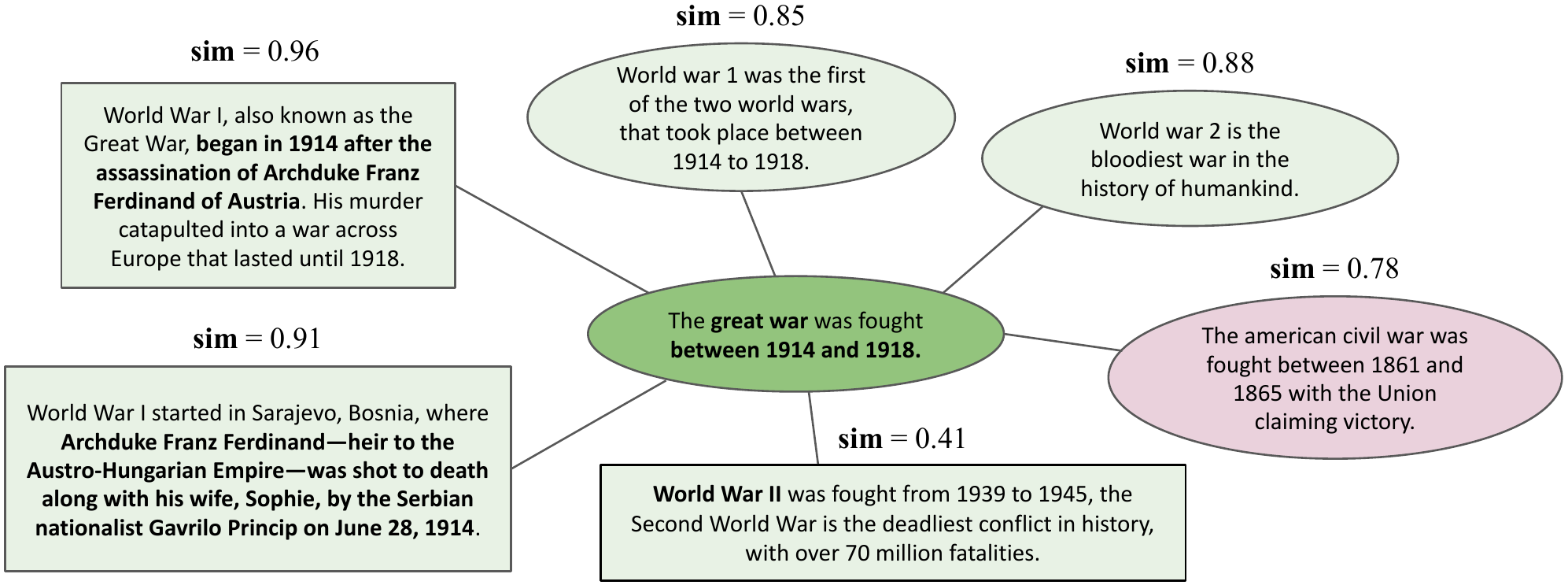}
    \caption{Shows an example inference output using IERL for a group of test sentences along with an anchor sentence $z$ ($z$ chosen for ease of illustration). The figure shows a group of instances shown in the oval and rectangular boxes (including $z$) and similarity measurements. For a pair of instances $z$ and one other instance from the group shown (let it be denoted by $z2$), we first find the closest sentences $x_1, x_2$ from the training set $X$ and compute two similarities as $s1 = \hat{BERT(x_1)} \cdot \hat{BERT(x_2)}$ and $s2 = \hat{ConceptNet(x_1)} \cdot \hat{ConceptNet(x_2)}$, where $\hat(.)$ represents normalizing the vectors as unit vectors. We display the greater of the two. The shapes are highlighted in green when the sum of the similarities is greater than or equal to $Models[x_1] \cdot Models[x_2]$, i.e., inference value = 1 (line 23 in algorithm \ref{algorithm}) and highlighted in pink otherwise, i.e., inference value = -1. The rectangular shape denotes the $s1 \geq s2$, and the oval shape denotes that  $s2 \geq s1$ (the parameter values also reflect the same in $\alpha_{x1}$ and $\alpha_{x2}$). Thus IERL is designed to provide a simple method to interpret the inference results for a group of test sentences.}
    \label{fig:alg_inference}
\end{figure}

Figure \ref{fig:alg_inference} shows an example inference output using IERL for a group of test sentences and an anchor sentence $z$ ($z$ chosen for ease of illustration). The figure shows a group of instances shown in the oval and rectangular boxes (including $z$) and similarity measurements. For a pair of instances $z$ and one other instance from the group shown (let it be denoted by $z2$), we first find the closest sentences $x_1, x_2$ from the training set $X$ and compute two similarities as $s1 = \hat{BERT(x_1)} \cdot \hat{BERT(x_2)}$ and $s2 = \hat{ConceptNet(x_1)} \cdot \hat{ConceptNet(x_2)}$, where $\hat(.)$ represents normalizing the vectors as unit vectors. We display the greater of the two. The shapes are highlighted in green when the sum of the similarities is greater than or equal to $Models[x_1] \cdot Models[x_2]$, i.e., inference value = 1 (line 23 in algorithm \ref{algorithm}) and highlighted in pink otherwise, i.e., inference value = -1. The rectangular shape denotes the $s1 \geq s2$, and the oval shape denotes that  $s2 \geq s1$ (the parameter values also reflect the same in $\alpha_{x1}$ and $\alpha_{x2}$). Thus IERL is designed to provide a simple method to interpret the inference results for a group of test sentences. 
\section{Conclusion and Future Work}
In this work, we propose Interpretable Ensemble Representation Learning (IERL) as an ensemble technique that demonstrates the interpretable combination of LLM and knowledge representations to result in a high-performance model that is robust to hallucinations and results in faster convergence in the number of optimization steps. Through our experiments, we see the promise of IERL as a method that advances research towards combining LLMs and knowledge graphs that retain both high performances and are interpretable by design (thus, addressing interpretability ambiguities during ablations and approximate post-hoc interpretations). In future work, we will explore different LLM and KG choices and vary the order of moments considered. Furthermore, we will explore other naturally interpretable combination functions (e.g., linear combination ensemble in this work) that can add layers of expressiveness to the interpretation (e.g., abstraction level in a hierarchy of concepts from a KG).

\section{Acknowledgement}

This work is built on prior work \cite{roy2023process,roy2023demo,roy2021depression,roy2021bknowledge,asawa2020covid,roy2022ksat,venkataramanan2023cook,roy2023knowledge,gaur2021can,roy2023proknow,tsakalidis2022overview,gupta2022learning,dolbir2021nlp,rawte2022tdlr,lokala2021edarktrends}, and supported by the National Science Foundation under Grant 2133842, “EAGER: Advancing Neuro-symbolic AI with Deep Knowledge-infused Learning" \cite{sheth2023neurosymbolic,sheth2021knowledge,sheth2022process}.
\bibliographystyle{unsrt}
\bibliography{references}
\end{document}